\pdfoutput=1
\documentclass[11pt]{article}

\usepackage[]{emnlp2021}
\usepackage{times}
\usepackage{latexsym}

\usepackage[T1]{fontenc}
\usepackage[utf8]{inputenc}

\usepackage{microtype}

\usepackage{amsmath}
\usepackage{mathtools}
\usepackage{amsthm}
\usepackage{amssymb}
\usepackage{enumitem}

\newcommand{\commentout}[1]{}

\usepackage{amsmath}
\usepackage{multirow}
\usepackage{booktabs}
\usepackage{graphicx}
\usepackage[normalem]{ulem}
\usepackage[cal=cm]{mathalfa}

\usepackage{amsthm}

\theoremstyle{definition}

\newcommand\jh[1]{\textcolor{blue}{[JH: #1]}}

\newcommand\nl[1]{{\it``#1''}}

\newcommand\geca{\textsc{Geca}}
\newcommand\ucoarse{\textsc{UAT}}
\newcommand\maxent{\textsc{CMaxEnt}}
\newcommand\random{\textsc{Uniform}}
\newcommand\baseline{\textsc{Baseline}}
\newcommand\sD{\mathcal{D}}

\newcommand\trainingset{\sD_{\emph{train}}}
\newcommand\testset{\sD_{\emph{test}}}
\newcommand\syntrainingset{\sD_{\emph{train}}^{\emph{Syn}}}
\newcommand\comptestset{\sD_{\emph{test}}^{\emph{Comp}}}

\newcommand\sampledset{\hat{\sD}_{\emph{train}}^{\emph{Syn}}}

\newcommand\nlschemaorg{\textsc{NL-SchemaOrg}}
\newcommand\synthschemaorg{\textsc{Syn-SchemaOrg}}

\newcommand\compem{\textsc{EM}_{\emph{comp}}}
\newcommand\iidem{\textsc{EM}_{\emph{iid}}}

\title{
\emph{Finding needles in a haystack}:\\ 
Sampling Structurally-diverse Training Sets from Synthetic Data for Compositional Generalization
} 
\author{Inbar Oren$^{1}$ ~~~~~ Jonathan Herzig$^{1}$ ~~~~~
Jonathan Berant$^{1,2}$ \\
$^1$School of Computer Science, Tel-Aviv University \\
$^2$Allen Institute for Artificial Intelligence \\
\small{\texttt{inbaroren@mail.tau.ac.il}}, \small{\texttt{\{jonathan.herzig,joberant\}@cs.tau.ac.il}}
}

\begin{document}
\maketitle

\begin{abstract}
Modern semantic parsers suffer from two principal limitations. First, training requires expensive collection of utterance-program pairs. Second, semantic parsers fail to generalize at test time to new compositions/structures that have not been observed during training. Recent research has shown that automatic generation of synthetic utterance-program pairs can alleviate the first problem, but its potential for the second has thus far been under-explored.
In this work, we investigate automatic generation of synthetic utterance-program pairs for improving compositional generalization in semantic parsing. Given a small training set of \emph{annotated} examples and an ``infinite'' pool of \emph{synthetic} examples, we select
a subset of synthetic examples that are structurally-diverse and use them to improve compositional generalization.
We evaluate our approach on a new split of the schema2QA dataset, and show that it leads to dramatic improvements in compositional generalization as well as moderate improvements in the traditional i.i.d setup. Moreover, structurally-diverse sampling achieves these improvements with as few as 5K examples, compared to 1M examples when sampling uniformly at random -- a 200x improvement in data efficiency.

\end{abstract}

\section{Introduction}
\label{sec:introduction}

Semantic parsers map natural language utterances to executable programs \cite{zelle1996learning,zettlemoyer05ccg}.
A worrying weakness of semantic parsers that has been recently exposed, is their inability to generalize at test time to new compositions \cite{finegan-dollak-etal-2018-improving,lake2018generalization, keysers2020measuring,kim-linzen-2020-cogs,gu2021beyond}. For example, a virtual assistant trained on the examples \nl{Show me Thai restaurants that allow pets} and \nl{How many hotels are there in Tokyo}, might not generalize to \nl{How many hotels in Tokyo allow pets?}. This type of out-of-domain generalization to new compositions constructed from components seen during training is commonly termed \emph{compositional generalization}.

\begin{figure}[t]
  \includegraphics[width=1\columnwidth]{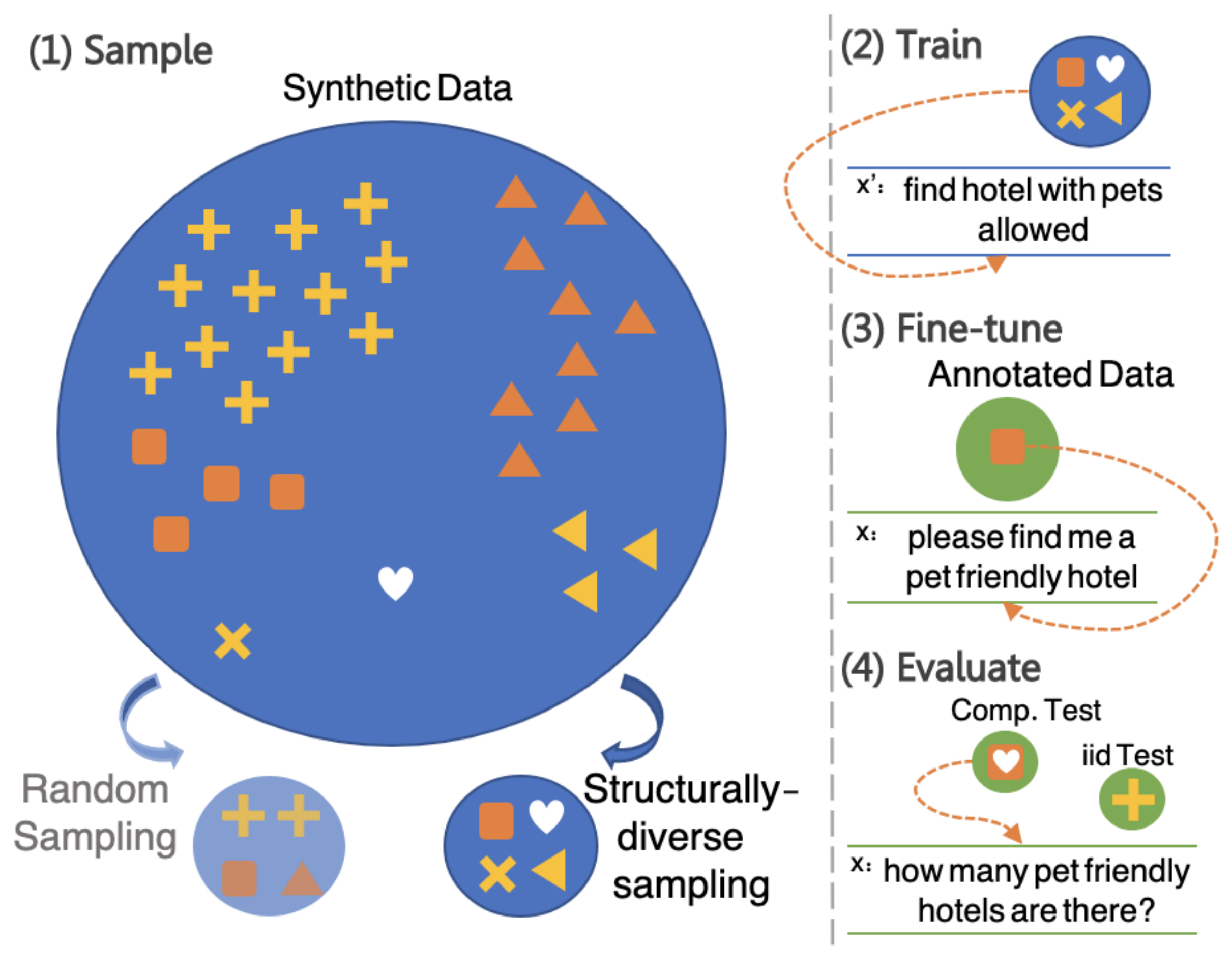}
  \caption{Given a small set of annotated data and a large pool of synthetic examples, we propose to sample a diverse training set of synthetic examples which includes a myriad of structures. Training a semantic parser on the synthetic data and fine-tuning on annotated data dramatically improves the parser's compositional generalization.
  }
  \label{fig:overview}
\end{figure}

Two high-level approaches have been considered for tackling compositional generalization: (a) designing models with a stronger compositional inductive bias  \cite{liu2020compositional,russin-etal-2020-compositional,zheng2020compositional,herzig2021span}, and (b) adding training data that will encourage a compositional solution \cite{akyurek2021learning,guo2020revisiting,wang2021learning,guo-etal-2020-sequence}. 
In the latter approach, typically a model is trained from labeled data \cite{jia-liang-2016-data,yu2020grappa,zhong-etal-2020-grounded}, and is used to later generate new examples. 
An alternative approach is to use a manually-built synchronous grammar that automatically generates programs paired with synthetic utterances \cite{wang2015building,cheng2018building,Weir2020DBPalAF}. Using a grammar allows generating large amounts of synthetic data that cover a wide range of program structures.
This has been shown to be useful in the i.i.d setup, and combined with paraphrase models, has led to high-accuracy parsers that are trained from synthetic data 
\emph{only} \cite{xu-etal-2020-autoqa}. 

In this work, we investigate the potential of using synthetic data, generated from a synchronous grammar, to improve compositional generalization.
\newcite{tsarkov2020cfq} have shown that large quantities of such data improve compositional generalization. However, they evaluated on synthetic utterances only, and did not examine generalization to natural language. Moreover, error rates were high in some compositional splits even when the training set was as large as 1M examples. In this work, we ask whether we can strategically sample a small and structurally-diverse training set and improve compositional generalization without incurring a high cost to training and consequently, to the environment \cite{Schwartz2020}. We hypothesize that a training set that encompasses a diverse set of structures can steer the model towards a compositional solution. 

We examine a realistic setup, where we have a small labeled training set  ($\sim$1,000 examples) and a large pool of synthetic utterances paired with programs (Fig.~\ref{fig:overview}), which are queries over a database (DB) in our setup.
We propose two methods for strategically sampling a diverse synthetic training set.
In the first, termed \emph{uniform abstract template (UAT)} sampling, we abstract queries by replacing DB constants with abstract tokens (e.g., replacing \texttt{petsAllowed} with \texttt{property}), and then skew the distribution towards  uniform sampling over the derived templates. 
This increases structure diversity in the training set, which intuitively should lead to better compositional generalization. In the second method, termed \emph{compound maximum entropy (CMaxEnt)} sampling, we consider the tree structure of every DB query, and following \newcite{tsarkov2020cfq} define \emph{compounds}, which are sub-trees in the query. We then heuristically solve an optimization problem, where our goal is to select a training set that has maximal entropy over compounds. This results in a training set with a diverse set of sub-structures, which should enhance compositional generalization.

We evaluate our approach on a new split of the Schema2QA dataset \cite{xu2020schema2qa}, in which it has been shown that synthetic data can lead to high accuracy parsers in the i.i.d setup \cite{xu-etal-2020-autoqa}.
We train an encoder-decoder model on synthetic data and subsequently fine-tune it on the small annotated data. We show that random sampling of synthetic data improves performance, where gradually increasing the size of the synthetic training set also improves compositional generalization. With 1M randomly-sampled examples, accuracy in the compositional split improves from 20.1$\rightarrow$37.7 and in the i.i.d split from 81.2$\rightarrow$85.0. When sampling structurally-diverse data, compositional generalization improves from 20.1 to $>$40 with as few as 5K examples, outperforming training with 1M synthetic examples. In addition, the i.i.d generalization is comparable to random sampling. \ucoarse{} and \maxent{} both lead to large improvements in compositional generalization, but \ucoarse{} is more effective than \maxent{}.

Overall, our work demonstrates that sampling diverse structures from synthetic data can lead to dramatic gains in compositional generalization at negligible cost, while preserving or improving performance in the i.i.d setup. Our code and data can be downloaded from \url{http://github.com/inbaroren/scfg-sampling-for-comp-gen}.

\begin{table*}[t]
\footnotesize
\begin{tabular}{lllllll}
\toprule
\boldmath$x$\textbf{:}&\multicolumn{6}{l}{\textit{show me a book with at least 2 awards .}} \\
\boldmath$x'$\textbf{:}&\multicolumn{6}{l}{\textit{which books have more than 2 awards}} \\
\boldmath$z$\textbf{:}&\multicolumn{6}{l}{\texttt{( @Book ) filter count ( award:Array(String) ) >= 2}}  \\

\midrule
\boldmath$x$\textbf{:}&\multicolumn{6}{l}{\textit{search for any books with a rating of 5 that also have 100 pages or more}} \\
\boldmath$x'$\textbf{:}&\multicolumn{6}{l}{\textit{what book gets number of pages at least 100 and gets the 5 mark ?}} \\
\boldmath$z$\textbf{:}&\multicolumn{6}{l}{\texttt{( @Book ) filter ratingValue:Number == 5 and numberOfPages:Number >= 100}}   \\ 
\midrule
\boldmath$x$\textbf{:}&\multicolumn{6}{l}{\textit{show me hotels with a fitness center}} \\
\boldmath$x'$\textbf{:}&\multicolumn{6}{l}{\textit{is there any hotels having fitness center in its amenity features}} \\
\boldmath$z$\textbf{:}&\multicolumn{6}{l}{\texttt{( @Hotel ) filter amenityFeature:Array(LocationFeatureSpecification) contains}}   \\
&\multicolumn{6}{l}{\texttt{''fitness center''}}   \\

\midrule
\boldmath$x$\textbf{:}&\multicolumn{6}{l}{\textit{can you find a hotel that accepts dogs ?}} \\
\boldmath$x'$\textbf{:}&\multicolumn{6}{l}{\textit{what hotels having pets allowed ?}} \\

\boldmath$z$\textbf{:}&  \texttt{( @Hotel )}& \texttt{filter}& \texttt{petsAllowed:} &\texttt{Boolean}& \texttt{==} & \texttt{true}   \\
\boldmath$z_\emph{abs}$\textbf{:}&  \texttt{( @table )}& \texttt{filter}& \texttt{property:} &\texttt{type}& \texttt{op} & \texttt{entity}   \\
\bottomrule
\end{tabular}
\caption{Examples from Schema2QA of utterance-query pairs ($x$,$z$) with their synthetic utterances ($x'$), in the books and hotels domains. In the bottom example, the abstract template ($z_\emph{abs}$) is included. Queries are in abbreviated syntax for clarity.}
\label{tab:dataset_examples}
\end{table*}

\section{Problem Setup}
\label{sec:setup}

We assume access to a small dataset of \emph{natural language} utterances paired with queries, $\trainingset=\{(x_i, z_i)\}_{i=1}^{N_{\emph{NL}}}$, and a large pool of \emph{synthetic} utterances paired with queries $\syntrainingset=\{(x'_i, z_i)\}_{i=1}^{N_{\emph{Syn}}}$, where $N_\emph{NL} \ll N_\emph{Syn}$. 
In this work, $\syntrainingset$ is generated with a synchronous context-free grammar, which provides wide coverage of query structures and tight control over the generated queries, but other methods of generating synthetic examples are possible \cite{andreas-2020-good,guo-etal-2020-sequence,wang2021learning}. Table~\ref{tab:dataset_examples} provides examples of natural language utterances, synthetic utterances, and queries in the ThingTalk language, a language designed for virtual assistants used in this work (through the Schema2QA dataset \cite{xu2020schema2qa,xu-etal-2020-autoqa}).

Our goal is to train a model using $\trainingset$ and $\syntrainingset$ and generalize to a test set $\testset$ sampled from the same distribution as $\trainingset$. More importantly, our model should generalize to a \emph{compositional test set}, $\comptestset$, which contains structures/compositions that are not observed in $\trainingset$ or $\syntrainingset$. We now describe this test set.

\paragraph{Compositional split}
\label{par:comp_split}
We define a compositional split following the \emph{abstract template split} proposed by \newcite{finegan-dollak-etal-2018-improving}, which is commonly used for assessing compositional generalization \cite{lee2019one,andreas-2020-good,Oren2020ImprovingCG}.
In this approach, queries are abstracted into templates that correspond to different structures. Templates are then partitioned into disjoint sets (train/development/test), which ensures that test time structures are never observed at training time. While prior work focused on abstracting entities only, by replacing any DB entity with the token \texttt{entity}, in this work we abstract queries into more coarse templates, e.g, table constants are replaced by the token \texttt{table}. Table~\ref{tab:abstraction_tokens} lists all abstracted query parts and their corresponding abstract token, and Table~\ref{tab:dataset_examples} (bottom) shows an example query $z$ and its abstract template $z_\emph{abs}$.
% where table constants are replaced by the \texttt{table} token, DB properties are abstracted to \texttt{property}, return types are abstracted to \texttt{type} and operators are abstracted to \texttt{op}.
Splitting with coarse templates increases the distance between train and test structures, which in turn increases the need for compositionality.\footnote{Preliminary experiments on synthetic data pointed that abstracting entities only might not lead to a compositional split that is challenging enough.}
Table~\ref{tab:split_examples} shows four examples, where the first two examples and the last two share an abstract template. In an \emph{i.i.d split} the first two (and last two) examples can appear in different sets, but in a \emph{compositional split} they must be either in the training set or test set, requiring compositional generalization.

\begin{table}[t]
\centering
\resizebox{1.0\columnwidth}{!}{
\begin{tabular}{lll}
\toprule
\textbf{Category} & \textbf{Token} & \textbf{Example} \\
\midrule
Entity & \texttt{entity} & \texttt{true}, \texttt{``Tokyo``} \\
Table & \texttt{@table} & \texttt{@org.schema.Book.Book} \\
Table property & \texttt{property} & \texttt{petsAllowed},\texttt{numberOfPages} \\
Entity or property type & \texttt{type} & \texttt{Array(String)}, \texttt{Number} \\
Operator & \texttt{op} & \texttt{>=}, \texttt{and}, \texttt{not}\\
Function & \texttt{func} & \texttt{count}, \texttt{sum}\\
Modifier & \texttt{func\_mod} & \texttt{asc}, \texttt{desc}\\

\bottomrule
\end{tabular}}
\caption{\label{tab:abstraction_tokens} 
List of all parts of the query we abstract and their corresponding abstract token.}
\end{table}

\begin{table}[t]
\centering
\resizebox{1.0\columnwidth}{!}{
\begin{tabular}{llccc}
\toprule
    &\multirow{2}{*}{\textbf{Example}} &   \textbf{iid} & \textbf{comp.}         \\  
                             &  & \textbf{split} & \textbf{split} \\ 

\midrule
\boldmath$x'$\textbf{:}&\textit{please search the hotels with pets allowed} & \multirow{2}{*}{train} & \multirow{2}{*}{train}\\
\boldmath$z$\textbf{:}&\texttt{( @Hotel ) filter patsAllowed:Boolean == true}   \\ 
\boldmath$z_{abs}$\textbf{:}&\texttt{( @table ) filter property:type op entity
}   \\ 
\midrule
\boldmath$x'$\textbf{:}&\textit{please search books with ebook format
} & \multirow{2}{*}{test} & \multirow{2}{*}{train}\\
\boldmath$z$\textbf{:}&\texttt{( @Book ) filter format:Enum == ebook
}   \\ 
\boldmath$z_{abs}$\textbf{:}&\texttt{( @table ) filter property:type op entity
}   \\ 
\midrule
\boldmath$x'$\textbf{:}&\textit{how many people are there
} & \multirow{2}{*}{train} & \multirow{2}{*}{test}\\
\boldmath$z$\textbf{:}&\texttt{aggregate count of ( @Person )
}   \\ 
\boldmath$z_{abs}$\textbf{:}&\texttt{func ( @table )
}   \\ 
\midrule
\boldmath$x'$\textbf{:}&\textit{how many hotels are there
} & \multirow{2}{*}{test} & \multirow{2}{*}{test}\\
\boldmath$z$\textbf{:}&\texttt{aggregate count of  ( @Hotel )
}   \\ 
\boldmath$z_{abs}$\textbf{:}&\texttt{func ( @table )
}   \\ 

\bottomrule
\end{tabular}}
\caption{\label{tab:split_examples} A compositional split prohibits the same abstract template to appear in both the training and test set, and hence tests compositional generalization. Above, examples 1-2 and 3-4 share the same template, so in an i.i.d split they can be assigned to different sets, while in a compositional split they must be either in the training or test set.}
\end{table}

\paragraph{Research questions}

Our experimental setup, where $\comptestset$ contains structures that are not observed in $\trainingset$ or $\syntrainingset$, allows us investigate several questions.
First, does training on the large synthetic dataset $\syntrainingset$ improve generalization to $\comptestset$ compared to training on $\trainingset$ only?
Second, if $\syntrainingset$ improves compositional generalization, can we make it more sample-efficient? Specifically, can we sample a smaller set $\sampledset$, such that $|\sampledset| \ll |\syntrainingset|$ and still improve compositional generalization. Last, can we do the above while preserving or improving generalization to the i.i.d test set, $\testset$? Answering these questions will be the focus of \S\ref{sec:methods} and \S\ref{sec:experiments}.

\section{Sampling a Structurally-diverse Training Set}
\label{sec:methods}

We first succinctly describe our model and training procedure (\S\ref{subsec:model}) and then turn to methods for sampling structurally-diverse training sets. 

\subsection{Model and Training}
\label{subsec:model}

In this work, we start from a pre-trained encoder-decoder model
\cite{lewis-etal-2020-bart}, as such models have been shown to provide a good initialization for fine-tuning semantic parsers \cite{furrer2020compositional,herzig2021unlocking}.
We then train our model in two steps \cite{yu2020grappa,wang2021learning}. First, on synthetic utterance-query pairs $(x', z) \in \sampledset$, and then on natural langauge utterance-query pairs $(x,z) \in \trainingset$. 
Training in two steps mitigates the gap in language variation between $\sampledset$ and $\trainingset$.
We train with the standard maximum-likelihood objective, maximizing the probability of the gold sequence, $z$.

\paragraph{Uniform sampling} Our baseline sampling method is to construct $\sampledset$ by sampling from  $\syntrainingset$ uniformly. This simulates sampling from the synchronous grammar directly, and can inform us whether synthetic data improves generalization to $\comptestset$, even if $\sampledset$ is very large.

\subsection{Uniform Abstract Template Sampling}
\label{ssec:methods_ucoarse}

We conjecture that a model is more likely to converge to a ``compositional solution'' if it observes at training time a multitude of different structures, and learns that sub-structures can occur in multiple contexts.
For example, if two properties always co-occur in the training set (e.g., \texttt{ratingValue} and \texttt{numberOfPages} in the second example of Table~\ref{tab:dataset_examples}), then the model might erroneously learn to always decode them together.

\begin{figure}[t]
  \includegraphics[width=1.0\columnwidth]{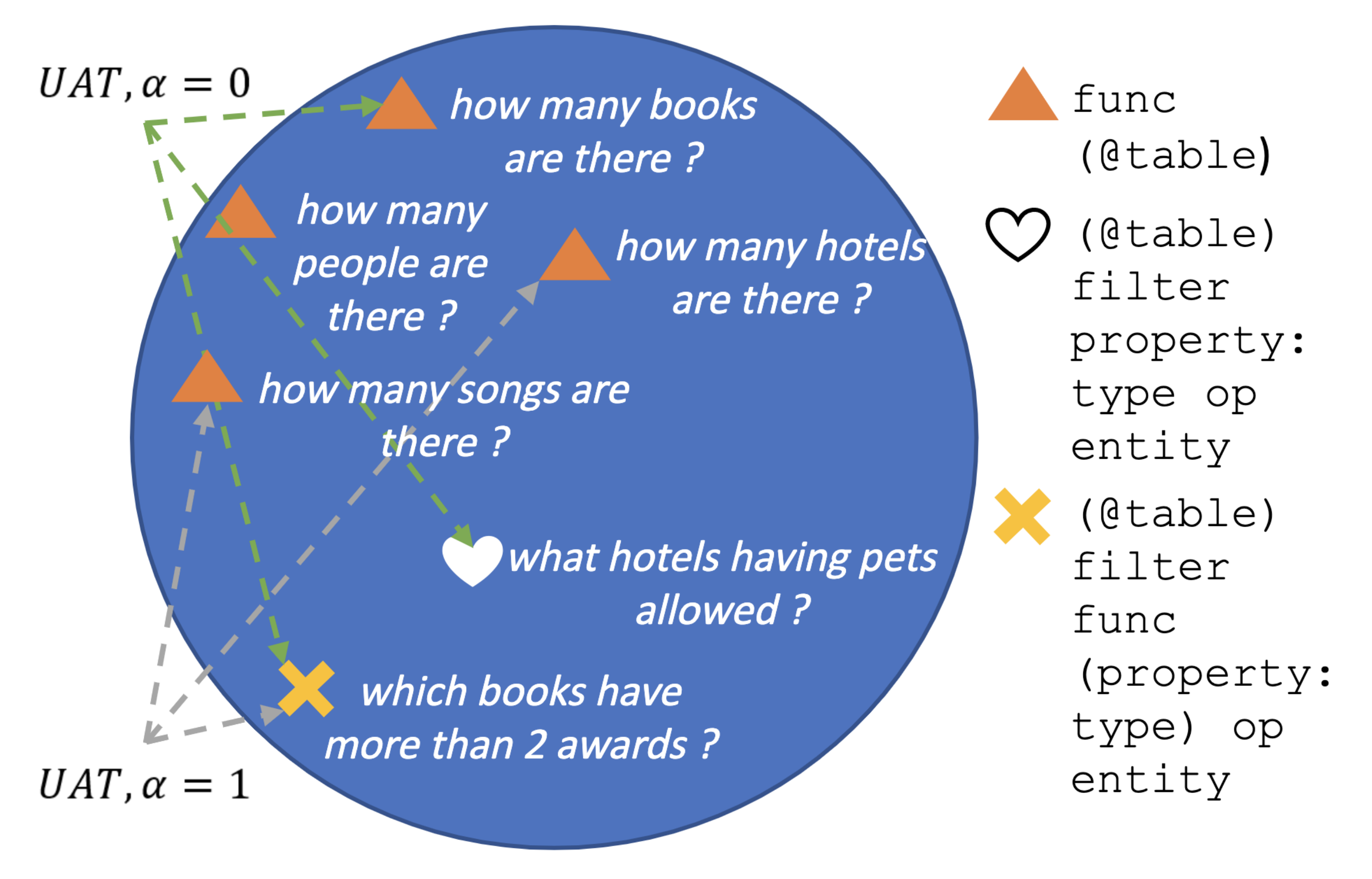}
  \caption{ \ucoarse{} sampling. When $\alpha=1$, \ucoarse{} is a uniform sample over $\syntrainingset$. As $\alpha \rightarrow 0$, the probability over abstract templates ($z_{\emph{abs}}$) becomes uniform. Consequently, the sample includes more abstract templates.
  }
  \label{fig:uaq_example}
\end{figure}

To achieve this goal, we define a sampling process, \ucoarse{}, that results in a more uniform distribution over abstract templates. In this process, we first sample an abstract template, and then sample an example conditioned on that template. We skew the distribution over templates to be close to uniform, which leads to templates that have few examples to be over-represented in the training set $\sampledset$. Thus, minimizing the loss over the training set will take into account a large number of templates, which should improve compositional generalization.  Typically, even if the number of examples in $\syntrainingset$ is large ($\sim$6M in our experiments), the number of abstract templates is much smaller (251 in our experiments, see \S\ref{sec:experiments}). Fig.~\ref{fig:uaq_example} illustrates this sampling process.

Formally, we construct $\sampledset$ by sampling from $\syntrainingset$ without replacement using the following procedure.
Let $\mathcal{T}$ be the set of templates in $\syntrainingset$, $T(z_i)$ be the abstract template for a query $z_i$, 
and $c(T(z_i))$ be the number of times $T(z_i)$ occurs in $\syntrainingset$. We estimate a probability distribution over templates, $p(T(z_i)) = \frac{c(T(z_i))}{|\syntrainingset|}$, and a distribution over examples conditioned on a particular template $u_{T(z_i)}((x'_i,z_i)) = \frac{1}{c(T(z_i))}$. Now we sample a synthetic utterance-query pair from the following distribution:
\begin{equation}
\label{eq:alpha}
q_{\alpha}((x'_i,z_i)) = 
\frac{p(T(z_i))^\alpha}{\sum_{T \in \mathcal{T}} p(T)^\alpha} \cdot u_{T(z_i)}((x'_i,z_i)),
\end{equation}
where $\alpha \in [0,1]$.
When $\alpha=1$, this corresponds to the aforementioned uniform sampling over examples (factorized over templates), but when $\alpha=0$, this corresponds to uniform sampling over the templates for which there still remain examples to be sampled. Values between $0$ and $1$ allow a smooth transition between uniform sampling over examples and over templates. In \S\ref{sec:experiments}, we will examine the effect of various values of $\alpha$ on compositional generalization for varying sizes of the sampled training set, $\sampledset$.

\subsection{Compound Maximum Entropy}
\label{ssec:methods_maxent}

\ucoarse{} sampling does not consider the similarity between different abstract templates, treating each template  independently. However, different templates potentially share some sub-structure that can be used for obtaining more diversity at the \emph{sub-structure} level.
We now consider \maxent{}, a method for sampling a synthetic training set with diverse sub-structures.

Recently, \newcite{keysers2020measuring} and \newcite{shaw2020compositional} used the notion of sub-structures, to construct a compositional test set. Given a program tree, they define \emph{atoms} to be nodes in the tree and \emph{compounds} to be sub-structures in the tree. Then, given a pool of examples, they partition it into two sets (train and test), such that the distribution over atoms is similar, but the distribution over compounds is different. 
Here, we adopt their definition of atoms and compounds, but for a different objective. We aim to sample a set $\sampledset$, such that the entropy over atoms and compounds is maximized. 
This will expose the model to a diverse set of atoms and compounds in multiple contexts, which should lead to compositional generalization. 

\begin{figure}[t]
  \includegraphics[width=1.0\columnwidth]{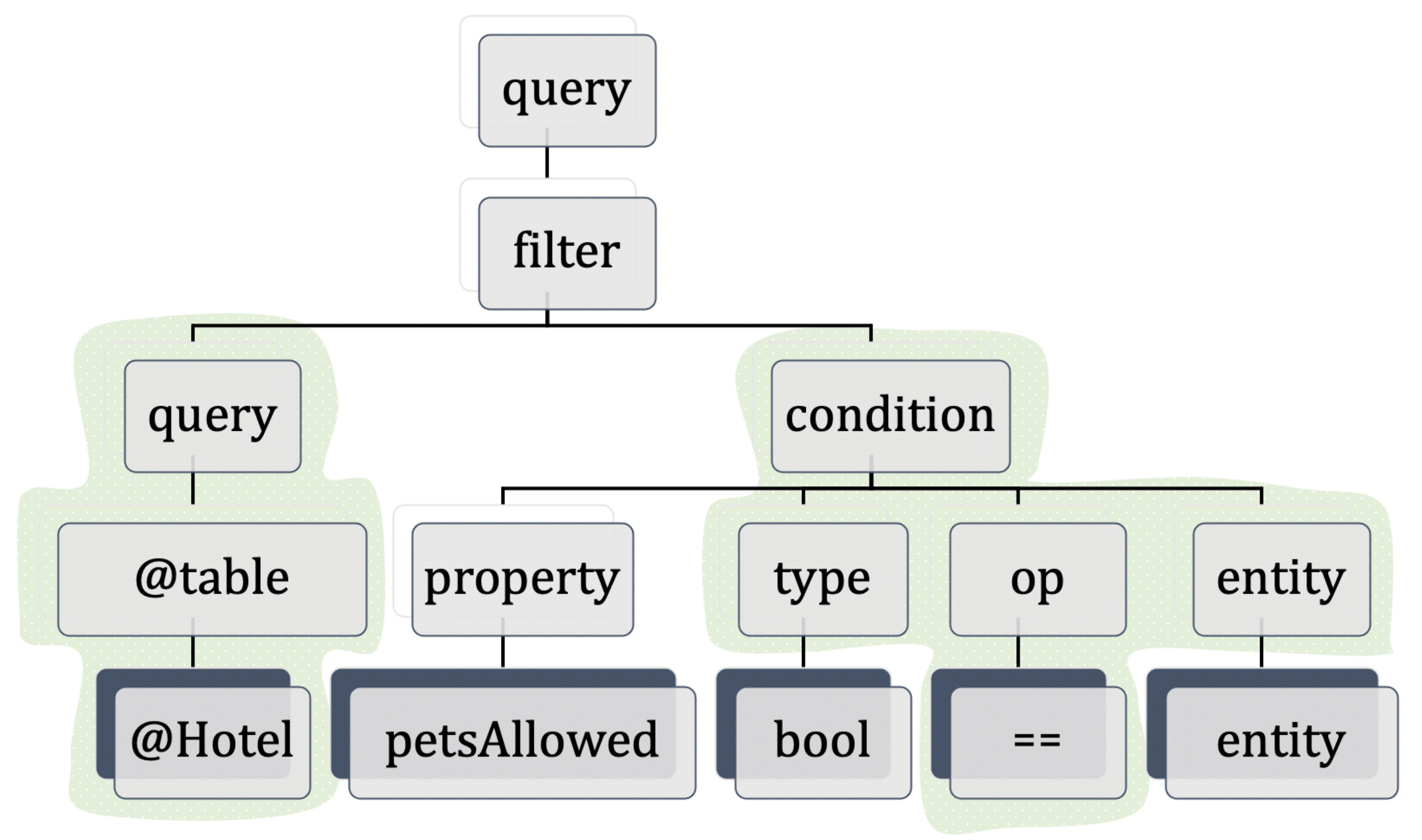}
  \caption{A ThingTalk parse tree (with abstract entities). Each node is an \textit{atom}, and any subgraph of height at most $2$ that has at least one terminal is a \textit{compound}. Two compounds are marked in green. 
  }
  \label{fig:parse_tree}
\end{figure}

\paragraph{Atom and compound distributions}
\label{par:compounds_def}
Queries in formal languages can be parsed into trees. We adopt the definition of \newcite{keysers2020measuring}, and define \emph{atoms} as any node in the tree, and \emph{compounds} as any tree of height $\leq 2$, that includes at least one tree terminal. We reduce the space of compounds by abstracting entity tokens (e.g., \texttt{''tokyo''}$\rightarrow$\texttt{entity}). Fig.~\ref{fig:parse_tree} shows an example tree with two compounds.

For a sampled set $\sampledset$, we use $p(a)$ to denote the frequency distribution of atoms in $\sampledset$, and $p(c)$ to denote the weighted frequency distribution of compounds in $\sampledset$. The compounds are weighted following \newcite{keysers2020measuring}, to avoid double-counting of compounds that mostly co-occur with their super-compounds.

\paragraph{Constructing $\sampledset$:}
\label{par:search_maxent}
Let $H(C) = - \sum_c p(c) \log p(c)$ be the entropy over compounds (and similarly for the atom entropy, $H(A)$).
Our goal is to find a synthetic training set, $\sampledset$ that maximizes $H(C) + H(A)$.

\commentout{
\begin{equation}
\begin{aligned}
    \max_{\sampledset} & \sum_{c\in \sampledset} -p(c)\log p(c) \\
    &+ \sum_{a\in \sampledset} -p(a)\log p(a) 
\end{aligned}
\end{equation}
Where $c$ and $a$ denote compound and atom respectively.
}

\begin{figure}[t]
  \includegraphics[width=1.0\columnwidth]{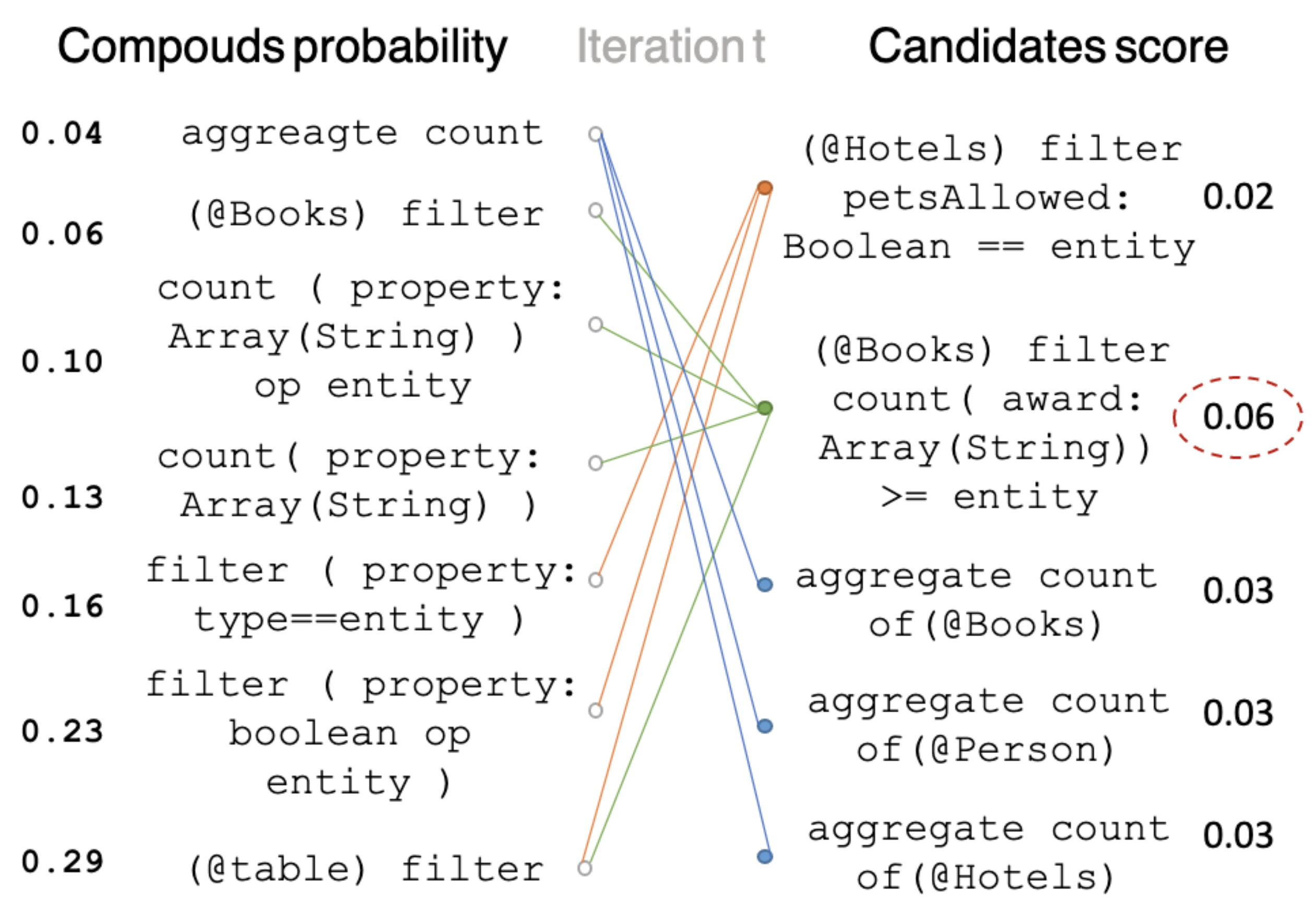}
  \caption{ One iteration in the \maxent{} sampling. On the left is a set of compounds and their probability $p(c)$ in the current training set. on the right is a list of candidate queries and the gain in compound entropy if they are selected. Colored lines show what compounds appear in what queries. Here, the second query is selected, and added to the sample with one of its accompanying utterances.  
  }
  \label{fig:maxent_samling}
\end{figure}

Finding the subset that maximizes the above objective is computationally hard, hence we use a greedy heuristic, termed \maxent{}, to approximate it. Despite its simplicity, we show in \S\ref{sec:experiments} that this approach improves entropy compared to a random sample. Specifically, we start with an empty training set, and in each iteration go over all examples in $\syntrainingset$ (with abstract entities),
choose the example that maximizes our objective, and add one of the corresponding non-abstract examples to $\sampledset$. We determine the number of iterations according to the desired target size of $\sampledset$. Fig.~\ref{fig:maxent_samling} illustrates a single iteration in this procedure.

\paragraph{Hybrid approach}
Our last approach combines our two sampling procedures, \ucoarse{} and \maxent{}. 
Specifically, in each iteration we sample an abstract template uniformly, and then use the greedy procedure for maximizing entropy over queries that correspond to the sampled abstract template. This results in a process that still improves the maximum entropy objective value but is skewed towards a uniform distribution over abstract templates.

\section{Experiments}
\label{sec:experiments}
We empirically evaluate the contribution of synthetic data to compositional generalization and the extent to which structurally-diverse sampling can improve data efficiency.

\subsection{Experimental Setting}

\paragraph{Data} 
Our work assumes access to a small annotated training set and a large pool of synthetic data, generated from a wide-coverage synchronous grammar. We build on work from \newcite{xu2020schema2qa}, who created the Schema2QA dataset, which contains natural language utterance-query pairs in the ThingTalk language for 6 domains of the \href{https://schema.org}{Schmea.org} onthology: \emph{restaurants, people, hotels, books, movies, and music}. Moreover, \newcite{xu-etal-2020-autoqa} presented AutoQA as part of the Genie toolkit\footnote{https://github.com/stanford-oval/genie-toolkit} for generating synthetic examples for Schema.org from a synchronous grammar \cite{campagna2019genie}. To obtain enough data we use the manually-labeled data from all 6 domains as our \emph{annotated data}, and term it \nlschemaorg{} (see \S\ref{ssec:error_analysis}). We generate 7M synthetic examples using AutoQA, and term it \synthschemaorg{}.

We construct an i.i.d split and a compositional split to \nlschemaorg{}, resulting in one common training set $\trainingset$, containing only 1,619 examples, a compositional development and test sets, and an i.i.d development and test sets (see Table~\ref{tab:split_stats} for statistics).
The training set, compositional development set, and compositional test set are all disjoint w.r.t their abstract template (see \S\ref{par:comp_split}). The i.i.d development and test sets are sampled from the same distribution as $\trainingset$. We create a compositional split of $\sampledset{}$, resulting in training/development/test sets that are disjoint in terms of their abstract template according to the templates in the compositional split of \nlschemaorg{}. We describe the exact procedure for splitting the data in Appendix~\ref{sec:supp_split_data_proc}.

\begin{table}[t]
\centering
\resizebox{1.0\columnwidth}{!}{
\begin{tabular}{lccc}
\toprule
\multirow{2}{*}{\textbf{Dataset}} & \multirow{2}{*}{\textbf{Split}}   & \textbf{\# examples}           & \textbf{\# new abstract templates} \\ 
                         &                          & (train / dev / test)  & (train / dev / test) \\
\midrule
\multirow{2}{*}{\textsc{Annotated}} & iid  & 1619  / 180 / 199      & 33 / 1  / 1  \\
                                   & Comp.  & 1619  / 188  / 304      & 33 / 19  / 20  \\ \midrule
\multirow{2}{*}{\textsc{Synthetic}}     & iid  & 5.8M / - / -     & 251 / - / - \\
                                   & Comp.  & 5.8M / 6K / 6K     & 251 / 11 / 10 \\

\bottomrule
\end{tabular}}
\caption{\label{tab:split_stats} Dataset statistics for the i.i.d split and compositional (comp.) split. \textit{\# new abstract templates} indicates the number of abstract templates unseen during training time for the development and test sets, and the total number of abstract templates for the training set.}
\end{table}

\commentout{
Our research requires access to natural langauge utterance-query  pairs and a synchronous context-free grammar on the same database. The work of \newcite{xu2020schema2qa} and \newcite{xu-etal-2020-autoqa} enables both on six schemas of Schema.org. For the natural langauge data we use the development set of the Schema2QA dataset \cite{xu2020schema2qa}, a semantic parsing dataset that maps utterances to ThingTalk queries in 6 domains (Table~\ref{tab:dataset_examples}). 
The Schema2QA development set was crowdsourced by showing a domain and a list of attributes to workers and asking them for natural language questions. 
The domain schemas are based on the Schema.org ontology, a commonly used representation for structured data in web pages. The domains are restaurants (extracted from Yelp), people (from LinkedIn), hotels (from the Hyatt hotel chain), books (from Goodreads), movies (from IMDb), and music (from Last.fm). 
We combine the data across domains to form a natural language dataset, ad denote it \nlschemaorg{}.

\paragraph{Synthetic Data Generation} To generate synthetic data, we use the AutoQA toolkit for Schema.org \cite{xu-etal-2020-autoqa}. The AutoQA algorithm relies on a comprehensive set of 800 domain-general templates \jh{can we use the terminology of a synchronous grammar instead? also, not sure we need all the details}, along automatically derived annotatios for each attribute in each domain. To synthesize data it uses the Genie algorithm \cite{campagna2019genie}, which iteratively expands the templates by substituting template slots with previously derived template parts. Since the expansion is exponentially, it randomly samples only a subset of derivations at each depth. By design, there's a large number of low depth programs, and relatively small number of high-depth programs.  
We generate $7M$ examples up to depth $10$ on the six schemas, and combine them to create a synthetic dataset, denoted \synthschemaorg{}.
}

\commentout{
\paragraph{Evaluation Splits} We first create a training/development/test splits of \nlschemaorg{} using both an iid split and a compositional split, so that both splits share the same training set. 
To achive that we first apply a compositional split on the data and keep the development and test sets. We then apply the iid split on the remaining data, obtaining iid development and test sets, and a train set which is valid for both splits.   
The compositional split of \synthschemaorg{} is created according to the \nlschemaorg{} compositional split, i.e, any synthetic example that its abstract template appear in \nlschemaorg{} compositional development/test set, is assigned to \synthschemaorg{} compositional development/test set. The rest of the data forms the training set . We also reduce the size of the synthetic compositional evaluation sets to $6k$ using random sampling. 
Table~\ref{tab:split_stats} presents exact statistics on the number of unique examples and abstract templates. 
}

\paragraph{Evaluation metric} We evaluate models using exact match accuracy, that is, whether the predicted query is identical to the gold query. We denote accuracy on the compositional and i.i.d splits as $\compem$ and $\iidem$ respectively. We report the average and standard deviation over 15 models, which are obtained by training on 3 different random samples $\sampledset{}$, each with 5 different random seeds.

In all experiments, we use $\iidem{}$ on the development set to determine early-stopping and for tuning batch size, learning rate and number of warmup steps (see hyper-parameter values in Appendix~\ref{sec:supp_hyper_parameters}).

\paragraph{Evaluated models} 
Our baseline parser is fine-tuned on the training set of \nlschemaorg{} (without pre-training), and is termed \baseline{}. We denote the rest of our experiments by the sampling method used to obtain $\sampledset$, where \random{} denotes uniform sampling, \ucoarse{} denotes abstract template sampling (\S\ref{ssec:methods_ucoarse}), \maxent{} denotes compound maximum entropy sampling (\S\ref{ssec:methods_maxent}). We denote the hybrid approach, combining the latter methods, as \maxent{}+\ucoarse{}. 

Importantly, we evaluate the effectiveness of our methods across different sizes of $\sampledset$. Overall, we are interested in the interactions between compositional generalization, i.i.d generalization, sampling method, and the size of the synthetic training set. We are especially interested in the effectiveness of our suggested sampling methods using smaller samples, hence limit the sample size to 120K.
We denote the size of $\sampledset$ by concatenating it to the model name, e.g., \ucoarse{}+5K corresponds to sampling with 5K synthetic examples.

As another baseline, we use \geca{} \cite{andreas-2020-good} as an alternative source for synthetic data. We use the publicly available code,\footnote{https://github.com/jacobandreas/geca} which takes the training set of \nlschemaorg{} and augments it with 1,342 new examples. We use these examples as $\sampledset$ in our setup. 

\subsection{Results}

\renewcommand{\arraystretch}{1.0}
\begin{table*}[t]
\centering
\footnotesize
\resizebox{\textwidth}{!}{
\begin{tabular}{l|l|c|ccccccc}
\toprule
\multirow{2}{*}{\textbf{Split}} & \multirow{2}{*}{\textbf{Method}}  & \multicolumn{8}{c}{\textbf{Sample Size}} \\  
                                            
                                                                    & &- &2K   &5K     &10K    &60K    &120K    &500K    &1M\\

\midrule
\multirow{6}{*}{\textit{comp.}}&\baseline{}& 20.1 $\pm$ 2.6 &   &     &   &    &    &    & \\
&\geca{}& 19.7 $\pm$ 1.0 &   &     &   &    &    &    & \\
&    \random{} &	 &           14.8 $\pm$	2.7&	22.2 $\pm$	2.4&	27.6 $\pm$	3&	28.6 $\pm$	4.1&	31.8 $\pm$	3.7&	35.8 $\pm$	1.3&	37.7 $\pm$	2.9 \\
   & \ucoarse{} &	   &     \textbf{27.8} $\pm$3.1&\textbf{41.4} $\pm$	3.3&\textbf{39.0}$\pm$2.4&	\textbf{43.1} $\pm$	6.5&	\textbf{43.0} $\pm$	4.7& &	\\			
   &  \maxent{} &	  &          17.0 $\pm$	2.8&	27.0 $\pm$	3.9&	31.8 $\pm$	1.8&	34.9 $\pm$	2.7&	40.2 $\pm$	1.6& & \\
    &\maxent{}+\ucoarse{} & &	21.1 $\pm$	4.8&	26.8 $\pm$	5.1&	34.9 $\pm$ 4.9&	39.1 $\pm$	5.2&	40.4 $\pm$	3.1& &	\\
\hline
\addlinespace[2pt]
\multirow{6}{*}{\textit{i.i.d}}&\baseline{}& 81.2 $\pm$ 3.2 &   &     &-   &    &    &    & \\
&\geca{}& 79.0 $\pm$ 2.1 &   &     &   &    &    &    & \\
&    \random{} &	  &          71.5 $\pm$	3.7&	78.9 $\pm$	4.4&	83.0 $\pm$	2.7&	84.4 $\pm$	2.5&	85.0 $\pm$	1.9&	\textbf{85.9} $\pm$	1.2&	85.0 $\pm$	1.3\\
    &\ucoarse{} &	    &  \textbf{80.7} $\pm$2.7&	\textbf{83.4} $\pm$	3.1&83.5 $\pm$	2.0&	\textbf{85.7} $\pm$	1.9&84.4 $\pm$	1.6&				& \\
    & \maxent{} &	   &         79.6 $\pm$	3.2&	81.1 $\pm$	4.1&	\textbf{83.8} $\pm$	1.2&	84.2 $\pm$	1.3&	84.6 $\pm$	2.2&				& \\
    &\maxent{}+\ucoarse{} &	&74.9 $\pm$	6.8&	78.7 $\pm$	4.4&	80.9 $\pm$	4.5&	85.1 $\pm$	1.1&	\textbf{85.9} $\pm$	1.8&				& \\

\bottomrule
\end{tabular}
}
\caption{\label{tab:main_verbose_results} Test compositional and i.i.d EM for all sampling methods (mean and standard deviation). Our structurally-diverse sampling methods allow us to use 200x less training data while improving $\compem$ significantly, and retaining comparable $\iidem$.}
\end{table*}

Table~\ref{tab:main_verbose_results} shows test results for both i.i.d and compositional generalization across all methods and synthetic training set sizes

\paragraph{Uniform sampling} 
Large amounts of synthetic data improve $\iidem$ on natural language data compared to \baseline{}, which agrees with the findings of \newcite{tsarkov2020cfq}.
Specifically, with 10K examples $\iidem$ improves 81.2$\rightarrow$83.0, and with 1M examples this improvement reaches 85.0 $\iidem$. As for compositional generalization, we observe improvements starting from 5K examples, with dramatic gains when training on 1M syntehtic examples -- 20.1$\rightarrow$37.7 $\compem$. To our knowledge, this is the first result showing that randomly sampling synthetic data from a wide-coverage grammar improves compositional generalization on natural language data.

When the size of $\sampledset$ is small, we observe a drop in $\iidem$, and also in $\compem$ when training with 2K examples. This shows that when $\sampledset$ is small, its distribution can potentially adversely affect training.

\begin{figure}[t]
  \includegraphics[width=1.0\columnwidth]{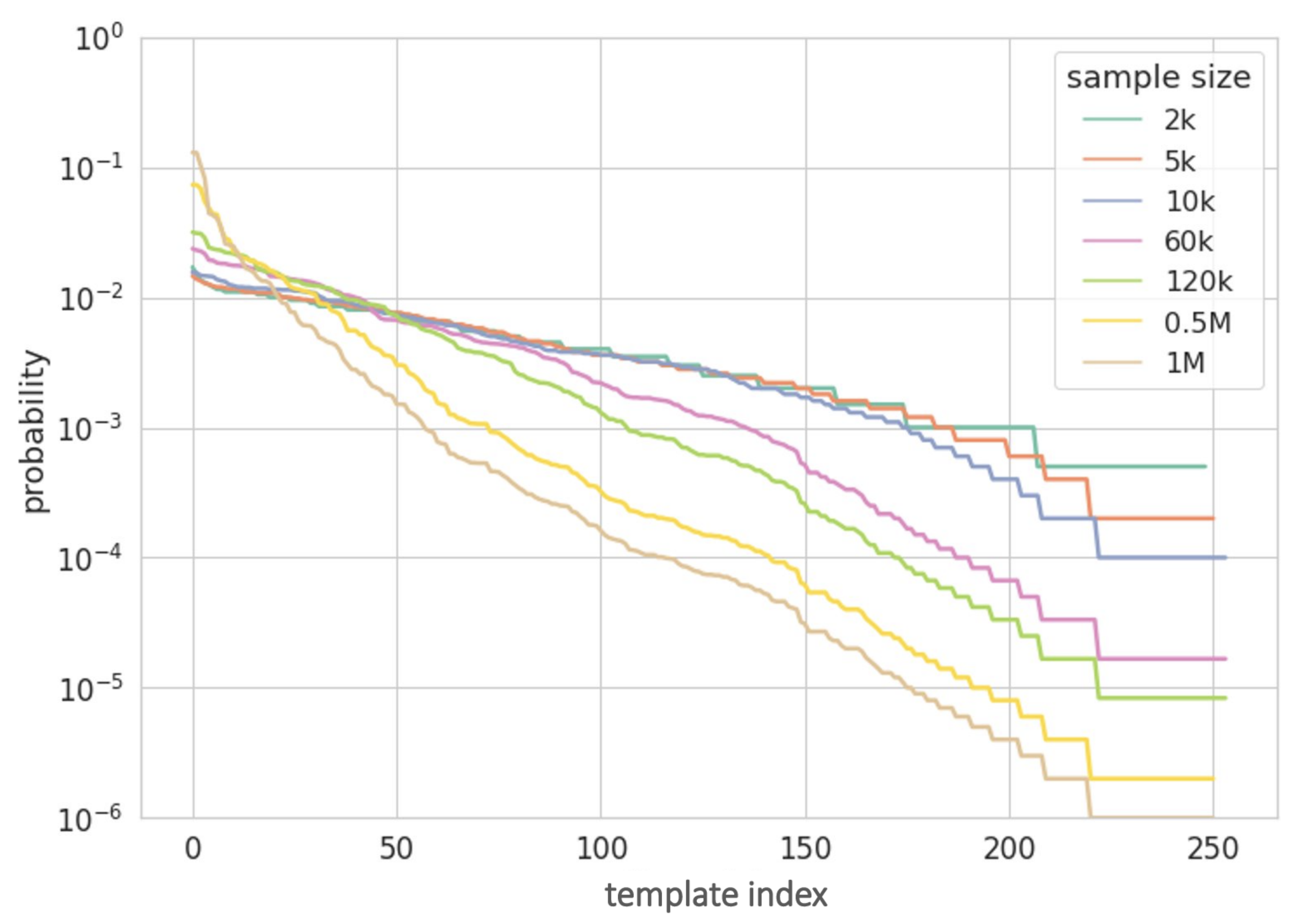}
  \caption{Distribution over abstract templates by sample size. The y-axis is in log-scale. The x-axis enumerates abstract templates sorted by frequency, i.e each point is a template. We observe that when the sample size is smaller, the distribution over abstract templates is more uniform.}
  \label{fig:abstract_query_freq}
\end{figure}

\paragraph{Abstract template sampling}
\ucoarse{} sampling dramatically improves compositional generalization even with very little synthetic data. With 2K examples $\compem$ improves from 20.1$\rightarrow$27.8, and with 5K examples $\compem$ is already at 41.4. This is a dramatic improvement compared to uniform sampling -- 3.7 $\compem$ points higher with \emph{200x less data}.

When further increasing the size of the synthetic data, improvement roughly plateaus, reaching 43.0 $\compem$ for 120K examples.
A possible explanation for this effect is that as the size of $\sampledset$ grows, the distribution over templates becomes more skewed, as shown in  Fig.~\ref{fig:abstract_query_freq}. 
Changing the composition of $\syntrainingset$ to contain more abstract templates by modifying the generation procedure in the AutoQA toolkit, and examining whether this leads to even greater gains in compositional generalization is an important question for future work.

\begin{figure}[t]
  \includegraphics[width=1.0\columnwidth]{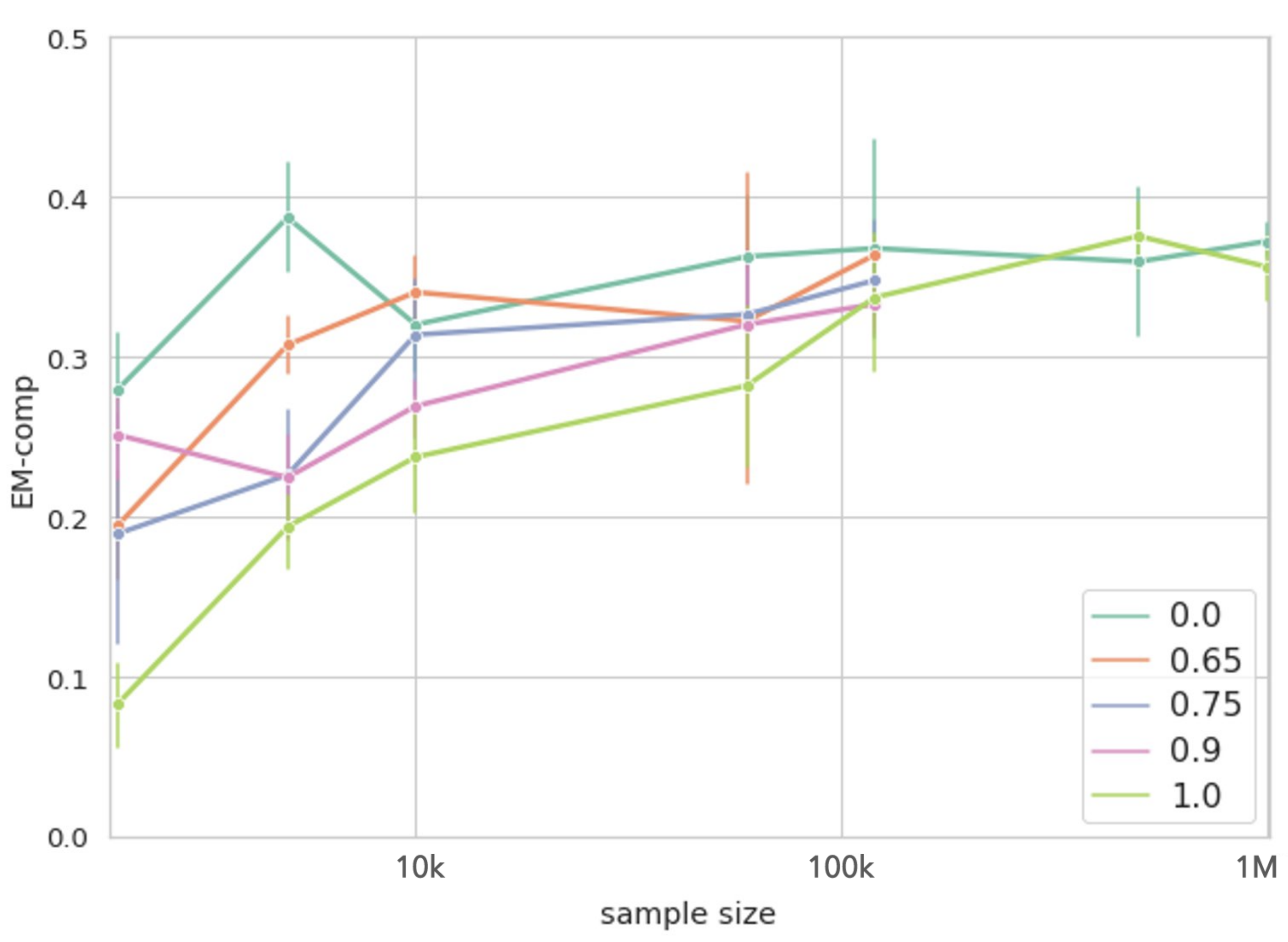}
  \caption{Accuracy on the compositional development set by size and $\alpha$ value. $\alpha=1$ is equivalent to a uniform sampling over examples, and as $\alpha$ decreases the distribution over abstract templates becomes closer to uniform. We report the average over 5 random seeds, and bars denote $95\%$ confidence intervals. x-axis is in log-scale.
  }
  \label{fig:comp_exp_results}
\end{figure}

To test if a smooth transition from $\alpha=1$ to $\alpha=0$ indeed leads to a smooth transition in compositional generalization, we train models with multiple values of $\alpha$.
Fig.~\ref{fig:comp_exp_results} confirms that 
tuning $\alpha$ from $1$ to $0$ yields a gradual improvement in $\compem$.

Last, $\iidem$ is comparable to $\random{}$, and even higher for smaller samples.

\paragraph{Compound maximum entropy}
\maxent{} improves compositional generalization with greater data efficiency compared to \random{}, as it improves $\compem$ at all sample sizes. With 120K examples, \maxent{} reaches 40.2 $\compem$, and surpasses \random{}+1M, at 37.7.

Still, \ucoarse{} outperforms \maxent{} in all cases. There are several possible explanations for this phenomenon. First, it might be the case that the distribution over abstract templates is the important factor in determining compositional generalization. In Appendix~\ref{sec:supp_samples_diversity} we show that indeed the distribution over abstract templates of \maxent{} is more skewed compared to \ucoarse{}. Second, our heuristic for optimizing entropy might be sub-optimal. While we do observe an entropy increase from 6.1$\rightarrow$7.1 in compound entropy, and from 3.9$\rightarrow$4.4 in atom entropy, it is possible that a better optimization procedure or a better definition of the notion of \emph{compound} will yield further gains in compositional generalization. We leave this investigation for future work.

\paragraph{Hybrid sampling} Combining 
\maxent{} and \ucoarse{} leads to improvements in $\compem$ over \maxent{} for all sample sizes (except 5K), but the overall performance does not surpass \ucoarse{}. 

\paragraph{\geca{}} Results are slightly lower than \baseline{}, $79.0$ $\iidem$ and $19.7$ $\compem$.
Sampling 1,342 examples from \geca{} is better than \random{}+2K, but worse than \ucoarse{}+2K. Thus, an advantage in the synchronous grammar is that we can easily sample a large number of synthetic examples that cover a wider range of structures.

\nlschemaorg{} comprises data from 6 different domains. In Table~\ref{tab:supp_domain_dev_result} in Appendix~\ref{sec:supp_dev_res}, we show development EM per domain. While the number of examples in each domain is small (a few dozen examples per domain), we still observe similar trends across all domains.

To summarize, our results suggest that abstract template diversity is an important factor in compositional generalization, and generating synthetic data with many abstract templates can dramatically improve compositional generalization.

\subsection{Analysis}
\label{ssec:error_analysis} 
We perform a manual analysis of models' predictions on \nlschemaorg{} compositional development set. We inspect 40 predictions of 8 models, and identify three major error types. The first type is \emph{structural} errors, which include syntax errors, misplacing parenthesis, and incorrect typing of properties. The second group is \emph{linking} errors, including, e.g., hallucination of entities, return of unsorted results, and missing an aggregation step. Third, we identify errors that are \emph{benign}, where the predicted queries are valid and equivalent to the target query. An example is using a wrong operator that does not change the query.
We also measure robustness to DB entity replacement in a query by grouping together any subset of development examples that only differ in a DB entity, and counting for how many groups all the examples are predicted correctly.

Table~\ref{tab:error_analysis} shows the results of the analysis. Our findings suggest two main benefits of larger $\sampledset$: (a) more frequently, the errors are benign, and (b) generalization is more robust to DB entity replacement. 
In addition, we find that using \ucoarse{} reduces structural errors, but increases linking errors, e.g, missing necessary \texttt{sort} or \texttt{filter} steps. Last, linking errors are the most common error type across all models.

Inspecting the predictions of \random{}+1M and \ucoarse{}+5K on the development set, we find that the abstract templates of correct predictions constitute roughly 40\% of the templates in the set, and are almost identical between the two models. We notice that ''hard'' templates are not necessarily longer or more nested.

\begin{table}[t]
\centering
\resizebox{1.0\columnwidth}{!}{
\begin{tabular}{l|ccc|c}
\toprule
\multirow{3}{*}{\textbf{Method}} & \multicolumn{3}{c}{(a)} &  (b) \\
& \textbf{Benign} & \textbf{Linking}   &  \textbf{Structural} & \textbf{Consistent} \\
&\textbf{Errors}&\textbf{Errors}&\textbf{Errors}&\textbf{Queries}\\

\midrule

\random{} \\
\;\;+5k&	7&	63&	30& 15\\
\;\;+60k&	8&	84&	8& 27\\
\;\;+120k&	10&	76&	14& 31\\
\;\;+1M&	11&	63&	26& 37\\
\maxent{}\\
\;\;+5k&	8&	62&	31& 24\\
\;\;+60k&	9&	52&	39& 31\\
\ucoarse{}\\
\;\;+5k&	8&	92&	0& 33\\
\;\;+60k&	0 &91&	9& 32\\

\bottomrule
\end{tabular}}
\caption{\label{tab:error_analysis} Error analysis. (a) a categorization of 40 predictions on the compositional development set, selected at random. Errors are partitioned by similar patterns to: \textbf{Benign errors}: the prediction is different but semantically equivalent to the target query, \textbf{Linking error}: a mismatch between information in input utterance and predicted query, and \textbf{Structural error}: wrong use of parenthesis and invalid queries. (b) the percentage of queries that are predicted correctly for any DB entity that occurs in the development set. }
\end{table}
\section{Related Work}
\label{sec:related_work}
 
\paragraph{Data augmentation}
Previous work studied different data augmentation techniques to improve i.i.d generalization in semantic parsing including synchronous grammars \cite{jia-liang-2016-data,yu2020grappa,xu-etal-2020-autoqa}, target side grammars  with neural generation models \cite{tran2020generating,wang2021learning}, and pre-training with auxiliary tasks \cite{yin-etal-2020-tabert,deng2021structure-grounded}. In the context of compositional generalization, data augmentation was achieved by re-combining training examples in new ways \cite{andreas-2020-good,akyurek2021learning}, or by back-translation \cite{guo2020revisiting}. Conversely, we generate data from an independent wide-coverage grammar and investigate data-efficient sampling through structured diversity.

Outside of semantic parsing, it has been shown in a grounded learning setup \cite{Hill2020Environmental} that increasing lexical diversity can improve out-of-distribution generalization.

\paragraph{Compositional Generalization}
In contrast to our work that focuses on sampling synthetic data, many other approaches have been suggested to improve compositional generalization in semantic parsing. These include new or modified model architectures \cite{li2019compositional,gordon2020permutation,guo2020hierarchical,Oren2020ImprovingCG,zheng2020compositional,herzig2021span,shaw2020compositional}, pre-trained language models \cite{furrer2020compositional}, intermediate representations \cite{herzig2021unlocking}, and meta learning \cite{lake2019compositional,conklin-etal-2021-meta}.

\paragraph{Data Selection} Our work is related to algorithmic approaches for reducing biases in datasets such as adversarial filtering \cite{le2020adversarial,sakaguchi2020winogrande} and representation debiasing \cite{li2019repair,li2018resound}. Our approach utilizes the structural nature of executable queries, and focuses on biases related to structural diversity.

\commentout{\section{Discussion and Related Work}
\label{sec:related_work}

Our work tackles the problem of compositional generalization through data augmenation, and we suggest sampling methods to increase data efficiency w.r.t the gains in compositional generalization. This section explores the connections in both aspects to related work. 

\paragraph{Data augmentation}
Previous work studied  data augmentation techniques to improve compositional generalization in semantic parsing, including re-combining training examples in new ways \cite{andreas-2020-good,akyurek2021learning}, or by back-translation \cite{guo2020revisiting}. Conversely, we generate data from an independent wide-coverage grammar, that was shown to improve i.i.d generalization \cite{xu-etal-2020-autoqa}. 

Other data augmentation techniques were shown to improve i.i.d generalization including target side grammars with neural generation models \cite{tran2020generating,wang2021learning}, and pre-training with auxiliary tasks \cite{yin-etal-2020-tabert,deng2021structure-grounded}. Our sampling methods can be applied to any synthesized data, however their effectiveness depends on its initial diversity.  
 
Outside of semantic parsing, it has been shown in a grounded learning setup \cite{Hill2020Environmental} that increasing lexical diversity can improve out-of-distribution generalization. 

\paragraph{Data Selection} Our work is related to adverserial filtering \cite{le2020adversarial,sakaguchi2020winogrande} and representation debias \cite{li2019repair,li2018resound}, algorithmic approaches to reduce biases which are hard to define in datasets. Our approach utilizes the structural nature of executable queries, and focuses on biases related to structural diversity. 
Moreover, to our knowledge these approaches were evaluated for classification tasks, and their generalization to generation tasks are under explored.
Since the structure of utterance-query pairs is harder to define, handling unkown biases in the data in addition to structural diversity can be beneficial. 

Despite our experiments showing that sampling diverse data can improve compositional generalization, compositional generalization remains a challenge. 
In \S\ref{sec:experiments} we propose how our methods can be further improved in the furture.
Moreover, the Genie toolkit, which we use for data synthesis, can be configured to new domains, allowing to extend our evaluation and understanding of the results.

We believe our findings can be generalized to other NLP tasks to improve data efficiency w.r.t both i.i.d and compositional generalization. This requires defining a structure over the training examples that can be used similar to the structures used in this work.

\paragraph{Compositional Generalization}
In contrast to our work that focuses on sampling synthetic data, many other approaches have been suggested to improve compositional generalization in semantic parsing. These include new or modified model architectures \cite{li2019compositional,gordon2020permutation,guo2020hierarchical,Oren2020ImprovingCG,zheng2020compositional,herzig2021span,shaw2020compositional}, pre-trained language models \cite{furrer2020compositional}, intermediate representations \cite{herzig2021unlocking}, and meta learning \cite{lake2019compositional,conklin-etal-2021-meta}.}

\section{Conclusion}
\label{sec:conclusion}
In this work, we for the first time explore whether generating large amounts of synthetic data from a synchronous grammar improves compositional generalization, and propose sampling methods that allow for more efficient training by generating structurally-diverse training sets.
We find that synthetic data dramatically improves generalization, and moderately improves i.i.d generalization, and that by uniformly sampling abstract templates, we can improve data efficiency by a factor of 200x.

In the past year, a myriad of approaches have been proposed for encouraging compositional generalization through modeling innovations, clever training procedures, and data augmentation techniques. Our work adds to the body of work that shows that data augmentation is an effective strategy even with small amounts of augmented data, when examples are carefully constructed. Moreover, data augmentation techniques can be easily combined with new models and training procedures, potentially leading to further gains in compositional generalization.

In addition, we believe our findings can be generalized to other NLP tasks to improve data efficiency w.r.t both i.i.d and compositional generalization. This requires defining a structure over the training examples that can be used similar to the structures used in this work.

\section*{Acknowledgements}
We thank Elad Segal, Ben Bogin, Matan Hasson and Ori Yoran for their useful suggestions. This research was supported in part by The Yandex Initiative for Machine Learning, and The European Research Council (ERC) under the European Union Horizons 2020 research and innovation programme (grant ERC DELPHI 802800). The second author was supported by a Google PhD
fellowship.

% \bibliography{all}
\bibliography{custom_rebiber_v2}
\bibliographystyle{acl_natbib}

\newpage

\appendix

\section{Data Split Procedure}
\label{sec:supp_split_data_proc}

To construct an i.i.d split and a compositional split we perform the following procedure. First, we construct a compositional split to \nlschemaorg{}, resulting in 3 sets that are disjoint in terms of their abstract template. We use 2 of them as the compositional development set and the compositional test set.
The third set is split i.i.d to trainining set, i.i.d development set, and i.i.d test set.
Next, we select the training, compositional development and compositional test sets from \synthschemaorg{}. We assign to each set the examples that their abstract template appear in the corresponding set of \nlschemaorg{}. Examples with a new abstract template (a template that do not occur in \nlschemaorg{}), are also included in the training set.  
% Next, we create a split of \synthschemaorg{} by assigning each example according to its abstract template's assignment in the split of \nlschemaorg{} (training/i.i.d dev/comp. dev/i.i.d test/comp. test). We add examples with a new abstract template to the training set. 
We downsample the compositional development and test sets of \synthschemaorg{} to 6K examples.

\section{Domain Statistics}
\label{sec:supp_domain_stats}
Table~\ref{tab:supp_domain_stats} shows the domain distribution in the synthetic and annotated datasets.

\begin{table}[t]
\centering
\resizebox{1.0\columnwidth}{!}{
\begin{tabular}{lccc}
\toprule
\textbf{Dataset} & \textbf{Domain} & \textbf{\# examples}  & \textbf{\# abstract templates} \\ 
\midrule
\multirow{6}{*}{\textsc{Annotated}}& books	& 356	&21\\
                                   & hotels	& 438	&24\\
                                    &movies	& 379	&23\\
                                    &music	& 310	&14\\
                                    &people	& 494	&27\\
                                    &restaurants&513&	44\\
                                   \midrule
\multirow{6}{*}{\textsc{Synthetic}}  & books&	1,022,222	&	107 \\
                                   & hotels	&477,759	&	151\\
                                   & movies	&1,101,019	&	112\\
                                   & music	&1,277,895&	108\\
                                  &  people&	1,339,201&		63\\
                                   & restaurants&	637,179	&	121\\

\bottomrule
\end{tabular}}
\caption{\label{tab:supp_domain_stats} Domain distribution in the synthetic and annotated datasets.}
\end{table}

\section{Training}
\label{sec:supp_hyper_parameters}
We implement and train our models using AllenNLP with PyTorch as backend, and initialize them using BART base. We conduct experiments on a machine with 8 NVIDIA GeForce GTX $2080$ GPUs and $40$ Intel(R) Xeon(R) Silver 4114 CPUs. The OS is Ubuntu $18.04.3$ LTS. 

\paragraph{Hyper-parameters} We use the Adam optimizer \cite{loshchilov2018decoupled} with learning rate selected from $\{0.00001,0.00002, 0.00003\}$. Batch size is selected from $\{1, 8\}$ for sample size $\leq$ 5000, and $\{24, 48, 64\}$ for larger samples.
We use a learning rate schedueler with polynomial decay and select warm up steps from $\{1000, 1500, 2000\}$ for sample sizes $\leq$ $1000$, and $\{2500, 3000, 3500, 4000\}$ for larger samples.
We use patience of $5$ epochs in pre-training, and $10$ epochs in fine-tuning. 
We use EM on the i.i.d development set as a metric for early stopping and selecting the best hyper-parameters. 
Hyper-parameters are fine-tuned for each sampling method and sample size separately on a single sample of \synthschemaorg{}. The patience is selected from $\{5,8,10\}$ when training on samples different than the one used for fine-tuning. Hyper-parameters for the models fine-tuned on \nlschemaorg{} are fine-tuned once on a \random{} sample.

\section{Sample Diversity}
\label{sec:supp_samples_diversity}
\begin{figure*}[t]
  \includegraphics[width=2.0\columnwidth]{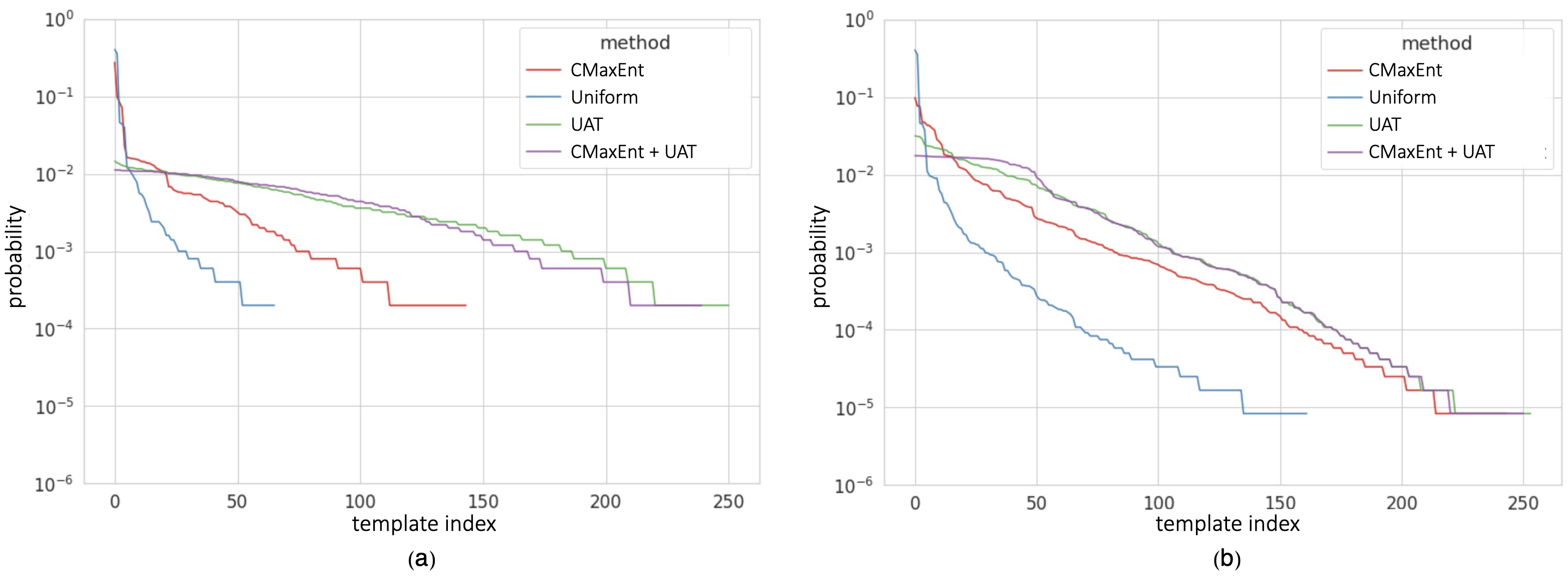}
  \caption{ Distribution over abstract templates by sampling method, for (a) a sample of size 5K, and (b) a sample of size 120K. The y-axis is in log-scale. The x-axis enumerates abstract templates sorted by probability, i.e each point is a template. 
  }
  \label{fig:templ_freq_5k_120k}
\end{figure*}

\begin{figure}[t]
  \includegraphics[width=1.0\columnwidth]{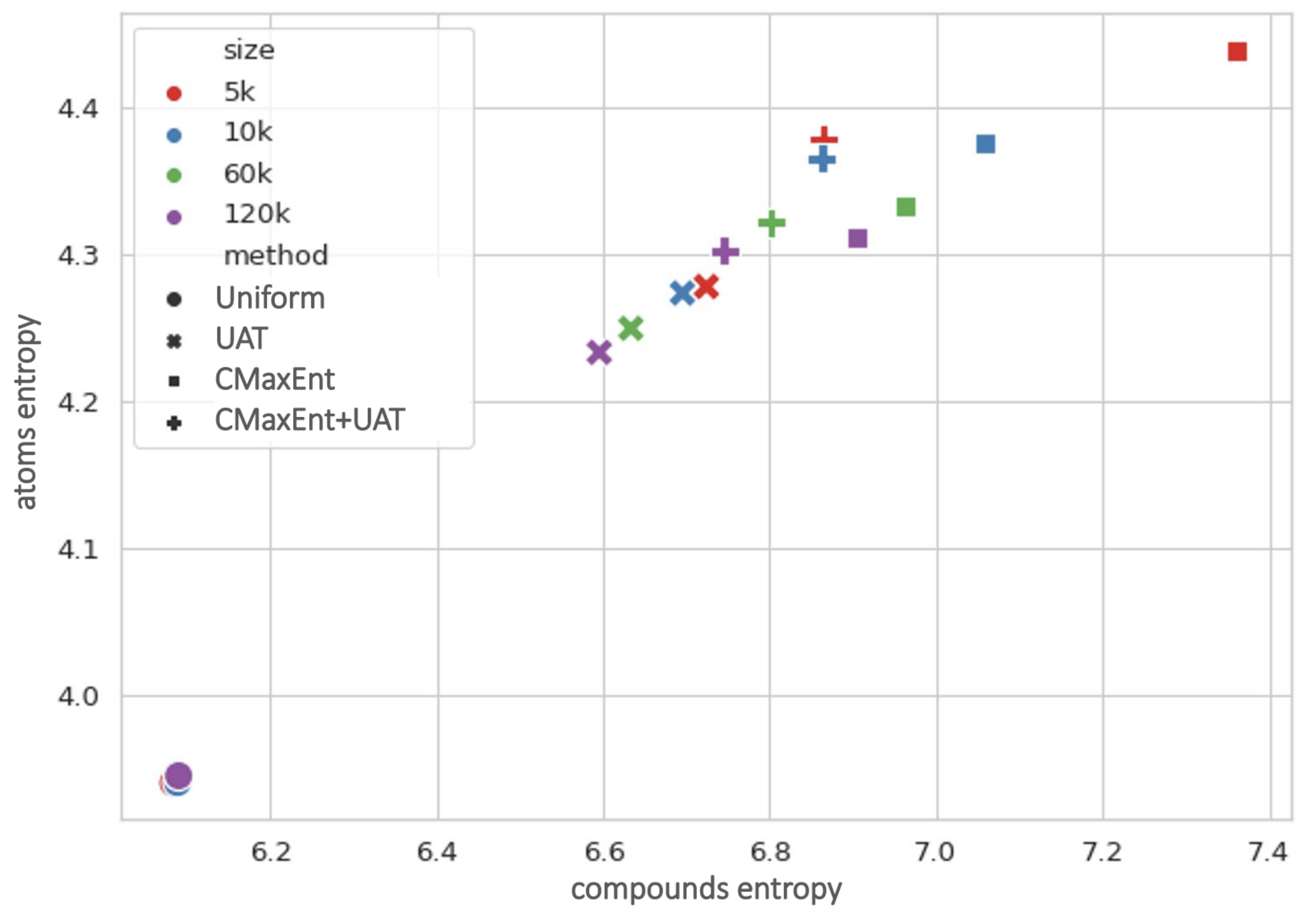}
  \caption{ Atoms and compounds entropy by sample size and sampling method, for one sample.
  }
  \label{fig:entropy_values}
\end{figure}

We measure structural-diversity in terms of abstract template distribution, atoms entropy and compounds entropy. Table~\ref{tab:coarse_templs_stats} compares the total number of abstract templates seen during both pre-training and fine-tuning between sampling methods and sample sizes.
Figure~\ref{fig:templ_freq_5k_120k} compares the normalized frequency of abstract templates between sampling methods for two sample sizes.
Figure~\ref{fig:entropy_values} compares the atoms and compounds entropy between sampling methods and sample sizes.
The above statistics are on a single sample from each method and size. 

\begin{table}[t]
\centering
\resizebox{1.0\columnwidth}{!}{
\begin{tabular}{lcccc}
\toprule
\multirow{2}{*}{\textbf{Size}} & \textbf{\random{}} & \textbf{\ucoarse{}}   &  \textbf{\maxent{}} & \textbf{\ucoarse{}} \\
&&&&\textbf{+\maxent{}}\\
\midrule
2k&	70	&256&	136&	192 \\
5k&	77	&258&	156	&247\\
10k&	95	&261&	214	&254\\
60k&	152	&261&	246	&258\\
120k&	169	&261&	252	&258\\
0.5M&	193&	&		&   \\
1M&223&	&	&\\

\bottomrule
\end{tabular}}
\caption{\label{tab:coarse_templs_stats} Number of abstract templates seen during pre-training and fine-tuning by size and sampling method.}
\end{table}

\section{Generalization of Synthetic vs. Annotated Data}
\label{sec:supp_results_synth_vs_annotated}
\begin{figure}[t]
  \includegraphics[width=1.0\columnwidth]{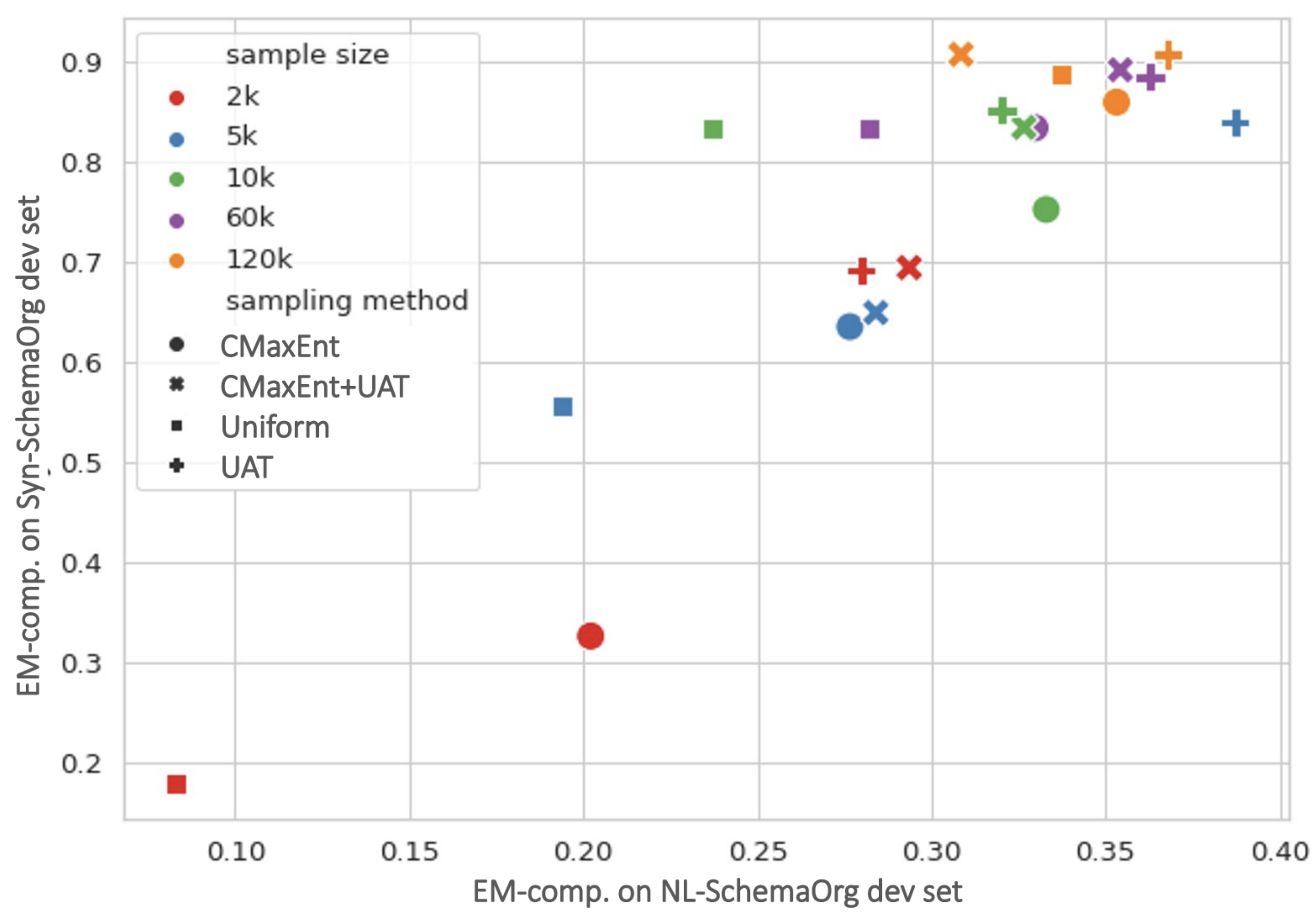}
  \caption{$\compem$ of the pre-trained model on \synthschemaorg{} development set, and the $\compem$ of the fine-tuned model on \nlschemaorg{} development set. We report the average over 5 random seeds on a single sample.
  }
  \label{fig:dev_scatter_results}
\end{figure}

Figure.~\ref{fig:dev_scatter_results} shows for each sampling method and sampling size, the $\compem$ of the pre-trained model on \synthschemaorg{} development set, and the $\compem$ of the fine-tuned model on \nlschemaorg{} development set. The relation between the performances is correlated for sample sizes smaller than 10K. We report the average over 5 random seeds and a single sample. 

\section{Development Results}
\label{sec:supp_dev_res}
Table~\ref{tab:supp_main_dev_results} contains the \nlschemaorg{} development set $\compem$ and $\iidem$ for all sampling methods and sampling sizes.
Table~\ref{tab:supp_domain_dev_result} shows the development set results by domain.

\begin{table*}[t]
\centering
\footnotesize
\resizebox{\textwidth}{!}{
\begin{tabular}{l|l|cccccccc}
\toprule
\multirow{2}{*}{\textbf{Split}} & \multirow{2}{*}{\textbf{Method}}&-  & \multicolumn{7}{c}{\textbf{Sample Size}} \\  
                                     & &  &2k   &5k     &10k    &60k    &120k    &500k    &1M\\

\midrule
\multirow{6}{*}{\textit{comp.}}&    
\baseline{} & 13.1\\
&\geca{} & 8.7 \\
&\random{}&	&	6.6&	17.0&	22.5&	30.7&	32.7&	36.7&	36.5\\
&\ucoarse{}&&	28.3&	36.1&	34.2&	35.2&	37.0&	&	 \\
&\maxent{} &&	19.5&	22.9&	32.8&	31.7&	32.5&	&	 \\
&\maxent{}+\ucoarse{}&&	19.6&	36.2& 31.8&	31.7&	32.3&		&      \\	
\hline
\addlinespace[2pt]
\multirow{6}{*}{\textit{i.i.d}}&   \baseline{} & 81.7\\
&\geca{} & 82.1\\
&\random{}&& 76.6&	82.1&	83.9&	86.0&	87.0&	88.8&	88.8 \\
&\ucoarse{}&& 81.6&	83.3&	85.1&	86.5&	87.8& &	 \\
&\maxent{} && 80.7 &	81.2&	85.0&	87.3&	87.8&	&	 \\
&\maxent{}+\ucoarse{}&&	78.4& 81.8 &	79.0&	87.2&	87.2&&\\

\bottomrule
\end{tabular}
}
\caption{\label{tab:supp_main_dev_results} Development EM for all sampling methods. We report average over 15 models, which are obtained by training on 3 different samples, each with 5 different random seeds.}
\end{table*}

\begin{table*}[t]
\centering
\footnotesize
\resizebox{\textwidth}{!}{
\begin{tabular}{l|l|c|l|cccccccc}
\toprule
\multirow{2}{*}{\textbf{Split}} &\multirow{2}{*}{\textbf{Domain}} &\multirow{2}{*}{\textbf{\#Examples}} & \multirow{2}{*}{\textbf{Method}}&   \multicolumn{7}{c}{\textbf{Sample Size}} \\  
                                     &&  & &2k   &5k     &10k    &60k    &120k    &500k    &1M\\

\midrule
\multirow{24}{*}{\textit{comp.}}&\multirow{4}{*}{Books}&\multirow{4}{*}{21}& 
\random{}& 16.2 & 21.9 & 21.9 & 32.4 & 39.7 & 40.9 & 34.1 \\
&&&\ucoarse{}& 31.4 & 47.6 & 35.2 & 48.6 & 40.5 &  &   \\
&&&\maxent{} & 20.9 & 30.5 & 43.8 & 29.5 & 39.1 &  &   \\
&&&\maxent{}+\ucoarse{}& 44.8 & 29.5 & 39.1 & 42.9 & 35.2 &  &   \\
\addlinespace[2pt]
&\multirow{4}{*}{Hotels}&\multirow{4}{*}{28}&    
\random{}& 2.9 & 15.7 & 24.3 & 26.4 & 35.7 & 50.0 & 50.6  \\
&&&\ucoarse{}& 27.9 & 46.4 & 39.3 & 47.1 & 41.1 &  &   \\
&&&\maxent{} & 21.4 & 26.4 & 41.4 & 40.7 & 42.1 &  &   \\
&&&\maxent{}+\ucoarse{}& 30.7 & 19.3 & 30.0 & 42.1 & 44.3 &  &   \\
\addlinespace[2pt]
&\multirow{4}{*}{Movies}&\multirow{4}{*}{33}& 
\random{}& 15.2 & 27.9 & 27.9 & 27.3 & 40.4 & 37.6 & 33.8  \\
&&&\ucoarse{}& 27.9 & 53.3 & 30.9 & 41.2 & 40.4 &  &   \\
&&&\maxent{} & 22.4 & 32.7 & 34.5 & 31.5 & 36.4 &  &   \\
&&&\maxent{}+\ucoarse{}& 31.5 & 32.7 & 35.1 & 38.2 & 32.7 &  &   \\
\addlinespace[2pt]
&\multirow{4}{*}{Music}&\multirow{4}{*}{31}&
\random{}& 0.0 & 0.7 & 7.1 & 13.6 & 13.4 & 11.6 & 11.8  \\
&&&\ucoarse{}& 4.5 & 17.4 & 5.2 & 9.7 & 15.0 &  &   \\
&&&\maxent{} & 6.5 & 11.0 & 9.0 & 10.3 & 11.0 &  &   \\
&&&\maxent{}+\ucoarse{}& 12.9 & 16.1 & 17.4 & 9.7 & 14.2 &  &   \\
\addlinespace[2pt]
&\multirow{4}{*}{People}&\multirow{4}{*}{23}&
\random{}& 6.1 & 9.6 & 6.1 & 9.6 & 15.2 & 10.4 & 9.4  \\
&&&\ucoarse{}& 13.9 & 15.7 & 15.7 & 12.2 & 13.0 &  &   \\
&&&\maxent{} & 8.7 & 7.0 & 9.6 & 8.7 & 10.4 &  &   \\
&&&\maxent{}+\ucoarse{}& 8.7 & 8.7 & 7.8 & 8.7 & 13.0 &  &   \\
\addlinespace[2pt]
&\multirow{4}{*}{Restaurants}&\multirow{4}{*}{52}&
\random{}& 9.6 & 30.4 & 39.2 & 45.0 & 49.7 & 57.3 & 49.4  \\
&&&\ucoarse{}& 46.9 & 44.6 & 50.8 & 48.9 & 54.2 &  &   \\
&&&\maxent{} & 31.1 & 43.1 & 48.9 & 55.4 & 55.0 &  &   \\
&&&\maxent{}+\ucoarse{}& 40.0 & 46.2 & 50.0 & 54.2 & 38.5 &  &   \\
\hline
\addlinespace[2pt]
\multirow{24}{*}{\textit{i.i.d}}& \multirow{4}{*}{Books}&\multirow{4}{*}{25}& 
\random{}& 79.2 & 83.2 & 84.0 & 83.2 & 87.3 & 88.0 & 88.7  \\
&&&\ucoarse{}& 82.4 & 84.8 & 86.4 & 86.4 & 88.0 &  &   \\
&&&\maxent{} & 83.2 & 85.6 & 84.8 & 86.4 & 87.2 &  &   \\
&&&\maxent{}+\ucoarse{}& 86.4 & 76.0 & 81.6 & 87.2 & 88.0 &  &   \\
\addlinespace[2pt]
&\multirow{4}{*}{Hotels}&\multirow{4}{*}{33}&
\random{}& 73.3 & 84.2 & 87.3 & 83.0 & 86.4 & 87.9 & 85.9  \\
&&&\ucoarse{}& 76.4 & 80.0 & 83.0 & 84.9 & 83.3 &  &   \\
&&&\maxent{} & 75.8 & 82.4 & 84.2 & 89.7 & 90.3 &  &  \\
&&&\maxent{}+\ucoarse{}& 83.0 & 78.2 & 72.7 & 87.9 & 87.9 &  &   \\
\addlinespace[2pt]
&\multirow{4}{*}{Movies}&\multirow{4}{*}{32}&
\random{}& 77.5 & 85.0 & 83.1 & 83.1 & 86.5 & 90.6 & 86.5  \\
&&&\ucoarse{}& 83.8 & 86.2 & 91.2 & 85.0 & 88.0 &  &   \\
&&&\maxent{} & 85.6 & 85.0 & 89.4 & 88.8 & 87.5 &  &  \\
&&&\maxent{}+\ucoarse{}& 83.1 & 74.4 & 88.1 & 89.4 & 90.0 &  &   \\
\addlinespace[2pt]
&\multirow{4}{*}{Music}&\multirow{4}{*}{12}&
\random{}& 71.7 & 78.3 & 85.0 & 81.7 & 80.6 & 85.0 & 91.7 \\
&&&\ucoarse{}& 73.3 & 85.0 & 88.3 & 81.7 & 87.5 &  &  \\
&&&\maxent{} & 81.7 & 85.0 & 91.7 & 86.7 & 88.3 &  &  \\
&&&\maxent{}+\ucoarse{}& 85.0 & 78.3 & 90.0 & 88.3 & 86.7 &  &   \\
\addlinespace[2pt]
&\multirow{4}{*}{People}&\multirow{4}{*}{45}&
\random{}& 84.4 & 89.8 & 88.4 & 87.6 & 90.0 & 88.4 & 90.7  \\
&&&\ucoarse{}& 86.2 & 88.0 & 85.8 & 88.0 & 90.0 &  &  \\
&&&\maxent{} & 84.9 & 90.7 & 85.8 & 90.7 & 89.8 &  &  \\
&&&\maxent{}+\ucoarse{}& 91.1 & 87.6 & 82.7 & 90.2 & 89.3 &  &   \\
\addlinespace[2pt]
&\multirow{4}{*}{Restaurants}&\multirow{4}{*}{33}&
\random{}& 80.0 & 87.9 & 79.4 & 85.5 & 87.9 & 92.1 & 88.9  \\
&&&\ucoarse{}& 76.4 & 83.6 & 84.9 & 81.2 & 88.4 &  &   \\
&&&\maxent{} & 80.0 & 80.6 & 84.2 & 85.5 & 89.1 &  &   \\
&&&\maxent{}+\ucoarse{}& 86.1 & 81.8 & 81.8 & 85.5 & 82.4 &  &   \\
\bottomrule
\end{tabular}
}
\caption{\label{tab:supp_domain_dev_result} Development EM for all sampling methods, by domain. We report average over 15 models, which are obtained by training on 3 different samples, each with 5 different random seeds. }
\end{table*}

\end{document}